\documentclass[a4paper,fleqn]{cas-dc}

\usepackage[natbibapa]{apacite}
\usepackage{enumitem,textcomp}
\usepackage{siunitx}

\usepackage{algpseudocode} 
\usepackage[ruled,vlined]{algorithm2e}
\SetKwInput{KwInput}{Input}                
\SetKwInput{KwOutput}{Output}              
\SetKwInput{KwParameter}{Parameter}

\usepackage{multirow}
\usepackage{graphicx}
\usepackage[normalem]{ulem}

\usepackage[activate]{microtype}
\def\tsc#1{\csdef{#1}{\textsc{\lowercase{#1}}\xspace}}
\tsc{WGM}
\tsc{QE}
\tsc{EP}
\tsc{PMS}
\tsc{BEC}
\tsc{DE}

\begin{document}

\let\WriteBookmarks\relax
\def\floatpagepagefraction{1}
\def\textpagefraction{.001}
\shorttitle{}
\shortauthors{Zhengxin Joseph Ye et~al.}

\title [mode = title]{Trading through Earnings Seasons using Self-Supervised Contrastive Representation Learning}                      

\author[1]{Zhengxin Joseph Ye}[type=editor]
\ead{z.ye18@imperial.ac.uk}
\address[1]{GLAM, Department of Computing, Imperial College London, UK}

\author[1]{Bj\"orn W.\ Schuller}[]
\ead{bjoern.schuller@imperial.ac.uk}

\begin{abstract}
Earnings release is a key economic event in the financial markets and crucial for predicting stock movements. Earnings data gives a glimpse into how a company is doing financially and can hint at where its stock might go next. However, the irregularity of its release cycle makes it a challenge to incorporate this data in a medium-frequency algorithmic trading model and the usefulness of this data fades fast after it is released, making it tough for models to stay accurate over time. Addressing this challenge, we introduce the Contrastive Earnings Transformer (CET) model, a self-supervised learning approach rooted in Contrastive Predictive Coding (CPC), aiming to optimise the utilisation of earnings data. To ascertain its effectiveness, we conduct a comparative study of CET against benchmark models across diverse sectors. Our research delves deep into the intricacies of stock data, evaluating how various models, and notably CET, handle the rapidly changing relevance of earnings data over time and over different sectors. The research outcomes shed light on CET's distinct advantage in extrapolating the inherent value of earnings data over time. Its foundation on CPC allows for a nuanced understanding, facilitating consistent stock predictions even as the earnings data ages. This finding about CET presents a fresh approach to better use earnings data in algorithmic trading for predicting stock price trends.

\end{abstract}



\maketitle

\section{Introduction}

The advent of high performance computing and the explosion of digital data in the late 1990s and early 2000s spurred the growth and popularity of machine learning into the specialised field of algorithmic trading, whether in supervised learning \citep{Chou2018}\citep{Tsai2019}\citep{Kumbure2022}, unsupervised learning \citep{Hu2018}, or reinforcement learning \citep{Huang2018}\citep{Minh2023}\citep{Ning2021}. 

We observe that research efforts in fusing machine learning with algorithmic trading often entail work on the modelling side \citep{Ye2023}\citep{Wang2019} and the data side \citep{Li2022}\citep{Soun2022}, with the research objectives often framed much more as a machine learning problem rather than a finance problem. Those are logical and legitimate research angles because financial time series data do carry embedded patterns available to be mined; auxiliary data such as news pieces and twitter feeds etc indeed offers added features and dimensions. However, from a practical standpoint, it is crucial to recognise that the stock market is primarily event driven and largely stimulated by the periodic injection of financial and economical data of various kinds at the macro level, such as U.S.\ nonfarm payrolls, Employment Situation Summary, and CPI data etc, \citep{Kurov2018}\citep{Heinlein2022}, as well as any data specific to individual companies, such as quarterly earnings releases \citep{Firth1976}. It is
the interpretation of such data by portfolio managers/traders that tips the balance of the underlying supply and demand of stocks which propel their prices to move toward certain directions. We believe it is imperative that a trading algorithm captures data release events and understands the jumps and changes in dynamics that follow \citep{barker2021}.

Post Earnings Announcement Drift (PEAD) as a stock market anomaly has been well studied by renowned economists from Ball and Brown \citep{Ball1968} to Fama and French \citep{Fama1993} and many others. It is a phenomenon when stock price continues to drift up for firms that are perceived to have reported good financial results for the preceding quarter and drift down for firms whose results have turned out worse than the market had anticipated. \citep{Ye2021}
investigated and illustrated the true predictability of PEAD based on a machine learning approach using a large dataset on earnings and financial metrics. Crucially, they highlighted the speed at which the market reacts to such information, causing the rapid vanishing of actionable trading signals. This led us to reconsider the prevailing practice in the literature: is relying primarily on stock price and volume time series, even when additionally augmented by other continual features such as twitter feeds and technical indicators \citep{Gao2016}, adequate enough to grasp the abrupt shifts in market dynamics during periods of heightened volatility, particularly when influenced by specific data types?
We argue that by relying on such continual data, we might miss out on capturing valuable predictive patterns during crucial market events. Therefore, to ensure a more holistic and responsive trading strategy, there is a compelling case for incorporating diverse datasets and employing advanced machine learning techniques to decipher the full spectrum of market behaviours. In this paper, we present an exploratory study into algorithmic trading through earnings seasons combining regular intra-day stock minutely-frequency data with irregular heterogeneous earnings data which drives the PEAD. We devise a novel algorithmic trading model, Contrastive Earnings Transformer (CET), which fuses data of various characteristics and granularity together through representations with the help of a Transformer \citep{Vaswani2017} and Contrastive Predictive Coding (CPC) \citep{Oord2018}.

Traditionally, Long Short-Term Memory (LSTM) and its variants featured heavily in algorithmic trading \citep{Dingli2017}\citep{Fischer2018} due to their natural fit with time series data, and this is particularly true when a single time series data is involved \citep{Sagheer2018}. However, as level of data complexity went up, LSTM's performance was found to be unsatisfactory and could not represent the complex features of sequential data efficiently, particularly if long interval multivariate time series with high non-linearity were involved \citep{LANGKVIST2014}. The emergence and proliferation of the Transformer model due to its groundbreaking architecture and remarkable performance across various domains quickly turned it into a more superior choice in place of LSTM in financial trading research \citep{Muhammad2023}\citep{Wang2018}. In this research, we employ a Transformer as an integral component of our model design and rely on its self-attention mechanism to facilitate robust feature extraction and foster contextual understanding of earnings data dynamics with raw price movements by attending simultaneously to different parts of the regular stock data time series and irregular earnings data, enabling it to model complex dependencies effectively.

It is widely known that machine learning models that rely on the backpropagation technique and optimization algorithms such as stochastic gradient descent (SGD) can suffer from random weight initialization which might not be optimal and can impact the initial behavior and convergence of the model during training \citep{Cao2017}. The Transformer model is not immune from it. In search of a remedy, our initial research found that previous works \citep{Haresamudram2019}\citep{Thomas2011}\citep{Saeed2019} had shown that, although the existence of intricate and infrequent movement patterns posed challenges for creating effective recognition models, integrating unsupervised learning into traditional pattern recognition systems led to promising outcomes, especially for feature extraction. Facing similar challenges with the diverse economic and price data and their inconsistent granularity and intensity, the proposed model adopts self-supervised pre-training to enable the model to learn powerful time-aware representations from the unlabeled price and earnings data. 

A catalogue of mainstream techniques has been developed and found successful for self-supervised pre-training, such as autoencoders and its variants \citep{Haresamudram2019}\citep{Varamin2018}, Masked Language Modelling (MLM), and contrastive learning, among a few others. MLM, which is used in BERT \citep{devlin2019}, is perhaps the most famous and has been used to obtain state-of-the-art results on a wide array of natural language processing (NLP) tasks. However, initial assessment of this study determines that Contrastive Predictive Coding (CPC) \citep{Oord2018}, a special type of contrastive learning model designed for sequential data, offers several advantages over MLM when working with time series data. Given the vast number of company stocks listed on major U.S.\ stock exchanges, our model needed to generalise well to unseen price sequences to be practically useful. MLM works better when the data is discrete (such as text data within a finite vocabulary), and its focus on predicting missing parts in masked positions may not explicitly encourage the model to capture broader temporal price patterns, potentially making it less robust in handling such time series prediction tasks. In contrast, CPC incorporates the notion of temporal context and learns the underlying patterns and structures by focusing on learning a useful representation of the data that maximises the similarity between the context vector and the future vector. This feature may enhance our model's ability to generalise and extrapolate its predictions to unseen stock time series and earnings data. Also, by focusing on predicting elements multiple time steps into the future, the model is encouraged to capture the underlying structure of the data that is less affected by noise or irrelevant variations. We have found this unique feature particularly important on a earnings release day when trading is typically most volatile.

In our exploration of algorithmic trading, we have taken a new approach to utilising earnings data by integrating CPC techniques. The task of melding the irregular release patterns of earnings data with high-frequency stock data posed significant challenges. The proposed solution is to turn to self-supervised learning. After a series of considered tests, we found that CPC emerged as a promising model to address our needs. The novel insights from this research include: (1) The CET model, through a series of experiments, showcases its potential in understanding earnings data, even when considering the variations across sectors or the diminishing relevance of the data over time. It offers a pathway for others to consider when dealing with irregular and vital data sources. (2) Our findings highlight the importance of earnings data in predicting stock movements. Even though its predictive potency may decrease as time progresses, its immediate impact on stock prices remains crucial. (3) One noteworthy observation was the adaptability of the CET model. As the influence of earnings data lessens over consecutive days, CET demonstrates an ability to adjust and refine its predictions, keeping pace with the evolving data. These discoveries enrich our understanding of algorithmic trading and pave the way for deeper investigations into the role of self-supervised models in financial decision-making.

\section{Related Work}

\subsection{Earnings Data and their impacts}
Ever since \citep{Ball1968} discovered Post Earnings Announcement Drift as a stock market anomaly in the 60s, quarterly financial earnings data have been playing an important role in various areas of financial analysis and forecast. Early machine learning-based studies by 
\citep{Olson2003} demonstrated the possibility of achieving excessive risk-adjusted returns when forecasting 12-month stock returns with a simple artificial neural network (ANN) using 61 financial ratios for 2352 Canadian stocks. Conducting fundamental and technical analysis together, both \citep{Hafezi2015} and \citep{Sheta2015} jointly utilised financial data and technical signals in trading company shares and equity index and reported satisfactory returns. \citep{Solberg2018} analysed 29\,330 different earnings call scripts between 2014 and 2017 using four different machine learning algorithms, and managed to achieve a low classification error rate and beat the S\&P500 benchmark in simulated trading. Similar research on earnings call transcripts by \citep{Medya2022} using a graph neural network reported reliable and accurate stock movement predictions and more importantly confirmed the overweighting of certain market-acknowledged variables such as Earnings-per-share (EPS). Both research studies demonstrated the predictive capabilities of auxiliary earnings information. Despite the success in both cases, we observe that their supervised learning's predictions were measured over days after earnings release. There was limited investigation into the immediate impact on intra-day price volatilities as market participants actively absorb and respond to the latest data that just came out. 

\subsection{Key Model Components}

One of the core tasks of algorithmic trading lies in financial time series prediction, which has enjoyed great success with the help of various deep learning technologies. As seen in the study by \citep{Zhang2023}, while LSTM remains a key component in constructing time series forecasting models such as in \citep{He2023}, recent trends show that advanced architectures like Transformers \citep{Xu2023}, GANs, GNNs, and DQNs are increasingly being utilized for price forecasting tasks, oftentimes showing exceptional capabilities in accuracy and convergence efficiency \citep{Kehinde2023}. This shift underscores how the financial industry is capitalizing on the latest developments in deep learning technologies.

With representation learning being proven to be an exceptionally valuable approach across many fields and applications, the Transformer model has emerged as a groundbreaking framework for this objective, with great success initially in natural language processing, as seen in \citep{Brown2020}\citep{devlin2019} which demonstrated that Transformer layers were able to overcome the performance of traditional NLP models. On the time series data front, \citep{Pérez2021} developed a Multi-Transformer model which merged several multi-head attention mechanisms to produce the final output and demonstrated that merging Multi-Transformer layers with other models led to more accurate stock volatility forecasting models. \citep{Zerveas2021} presented a transformer encoder-based multivariate time series representation learning framework with a focus on unsupervised pre-training, which was proven very successful extracting vector representations of unlabelled data. This particular result inspired us to introduce a pre-training phase to our model, except that we have specifically chosen self-supervised pre-training whose key idea is to employ a `pretext' task where the model is asked to predict some part of the data given the rest, creating some sort of `supervised signal' in the process which benefits downstream tasks. In fact, powerful Transformer-based models with self-supervised learning capabilities have been developed, such as GPT \citep{Radford2018} and BERT \citep{devlin2019}.

One very effective family of self-supervised representation learning is contrastive learning, based on which a lot of successful models have been developed, such as SimCLR \citep{Chen2020} for visual representations, TimeCLR \citep{Yang2022} for time series, and CPC \citep{Oord2018} whose learnt representations have been found to deliver strong performance in numerous applications. For example, \citep{Baevski2020} utilise CPC on raw speech signals from a large unlabelled corpus to improve speech recognition performance on smaller labelled datasets. \citep{Lowe2019} used the CPC's InfoNCE loss to calculate the loss inside the module by dividing the deep neural network into a set of gradient isolation modules. \citep{Wang2021} introduced a copula-based contrastive predictive coding (Co-CPC) method in order to address the issue of inadequate generalisation caused by uncertainty in data and models. Co-CPC considers the dependencies between a stock class, sector, and related macroeconomic variables, and learns stock representations from a micro perspective in a self-supervised manner. This allows for the mapping of stock characteristics to a generalised embedding space. \citep{Eldele2021} proposed the TS-TCC model which would treat the data with two different augmentations as a pair of positive samples, then map the augmented data to the latent space through a convolutional layer, and additionally learn the feature from latent space through the Transformer module.

\section{\label{sec:models and methods}Models and Methods}

\begin{figure*}[t!]
  \centering
  \includegraphics[width=0.8\textwidth]{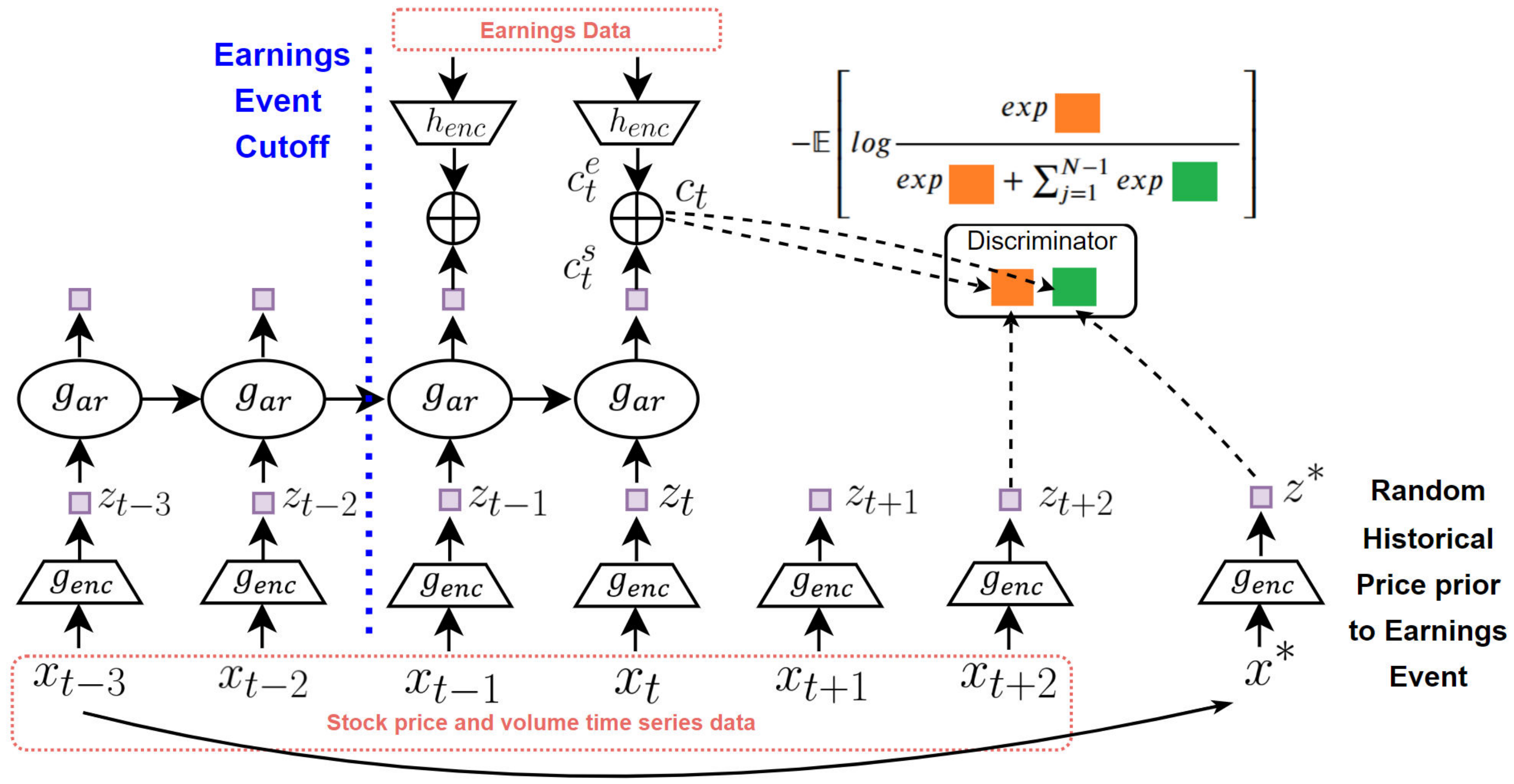}
  \caption{Representation learning of the combined price, volume, and earnings data via CPC. Price and volume data, as well as earnings data, are separately encoded and combined through element-wise addition to form the context vector $C_{t}$ whose mutual information with future price and volume data representations, $z_{t+k}$, are optimised.}
  \label{fig:model_structure}
\end{figure*}

Self-supervised representation learning based on Contrastive Predictive Coding (CPC) plays a central role in allowing the CET model to digest important earnings data while reading in historic price and volume data series on a per-minute basis. Inituitively a CPC approach involves three steps: First, the approach involves condensing complex, high-dimensional data into a simpler, smaller latent space, making it easier to predict outcomes. Next, advanced autoregressive model is used within this simplified latent space for forecasting into the future. Lastly, the model’s training is enhanced end-to-end using Noise-Contrastive Estimation (NCE) in the loss function. Fundamentally, CPC is designed to capture and learn the essential information shared across various segments of a complex signal while filtering out less important details and noise. In modeling time series and high-dimensional data, methods that typically predict the immediate next step utilize the signal's local continuity. However, for predictions further into the future, a model needs to identify broader, more global structures, as the shared information decreases with time. This approach focuses on 'slow features' that are relevant over extended time periods \citep{Oord2018}, which are typically more insightful for understanding the data's underlying patterns, and especially so on the extended impact by an event on time series data, which is exactly what we are modelling for earnings release. This is achieved by employing a "pretext task" which typically involves predicting some part of the input data from other parts, which helps the model learn useful representations of the data. Specifically, the pretext task involves predicting future segments of a signal based on past segments. This helps the model capture and understand the underlying structure and features of the data, which can then be used for various downstream tasks such as classification.

With those model principles in mind, the network structure of the CET model as well as the optimization goal of the pretext task are depicted in figure \ref{fig:model_structure}. In the initial phase, the process involves a pre-training stage comprising two separate data encoding activities. To learn the intra-day price context $c_{t}^{s}$ a non-linear encoder $g_{enc}$ is first used to map the input sequence of observed price and volume data $x_{t}$ to a sequence of latent representations $z_{t}=g_{enc}(x_{t})$. While \citep{Oord2018} studied the choice of various convolutional neural networks including a 1D CNN and ResNet in their initial research, a Multilayer Perceptron (MLP) with one hidden layer and a linear activation at the output layer is deemed sufficient for the task of price data embedding. A Transformer $g_{ar}$, which is the choice of the auto-regressive model, then summarizes all $z_{t}$ in the latent space to produce a stock context representation $c_{t}^{s}=g_{ar}(z_{\leqslant t})$. The earnings data representation $c_{t}^{e}$ is generated through a regular Autoencoder which is combined with $c_{t}^{s}$. This whole process generates effective data representations, aggregating time series data and periodic data into a single context vector representation $c_{t}$. These representations are then employed in a preliminary training activity, i.e. the pretext task, in the framework of CPC.

Once all of the network weights have been learned in the unsupervised learning pretext task, the model will be further optimized through a fine-tuning phase which involves predicting the price level in the next minute on the learnt representation. Actual price levels are used in this step since the entire network's weights will be adjusted in a supervised-learning style training routine. This step will complete the self-supervised learning phase and the weights from $g_{enc}$, $g_{ar}$, and $h_{enc}$ will then be frozen for the actual stock trading experiments. To do that, another MLP with one hidden layer is used whose neuron weights will be optimized to predict stock price gain/loss in percentage terms at $t+1$ with respect to the price at $t$.

\subsection{\label{sec:level2}Encoding of Earnings Data}
we propose to encode the diverse and heterogeneous set of earnings data into a compact vector representation using an autoencoder \citep{Charte2020}, reducing the dimensionality of the feature space but most importantly allowing the earnings data to be aligned and connected with price and volume data encoded through the Transformer. The autoencoder consists of an encoder network $E(x)$ and a decoder network $D(c)$. The encoder network takes as input the group of pre-processed earnings features $x_{e} \in \mathbb{R} ^{D}$ with $D$ being the earnings data dimension and maps them to a $d$-dimensional context vector $c_{t}^{e}$. The decoder network then reconstructs the original input from the $d$-dimensional latent space representation using a reverse network structure and calculating the reconstructed output $x' = D(E(x))$. The decoder minimises the reconstruction error between $x_{e}$ and $x^{'}$ through mean squared error loss $\frac{1}{N} \sum_{i=1}^{N} \left\|x_{e;i} - x_i'\right\|^2$, learning a compressed representation of the input earnings $c_{t}^{e}$ and capturing the most salient features of the data while discarding noise and less important details.  


It is noteworthy to mention that we have assessed the suitability of using a Variational Autoencoder (VAE) \citep{Gündüz2021}. We determined that capturing underlying probabilistic structures or generating new samples was of lesser importance. Instead a regular autoencoder has been chosen, due to its simpler architecture and more straightforward training process when compared to VAEs. This choice is driven by the priority for high-fidelity reconstruction of input data. Through the direct minimization of reconstruction loss, the autoencoder aims to accurately replicate the original input, a key factor in our decision to use this model.

\subsection{\label{sec:encoding_of_data}Encoding of Price and Volume data}
Our stock data at every minute includes the closing stock price and average trading volume within the minute time frame. In particular, each training sample of price and volume data is $X \in \mathbb{R} ^{\omega \times \upsilon }$ which is a time series of length $\omega$ with $\upsilon$ feature vectors, $x_{t} \in \mathbb{X}^{\upsilon}:X \in \mathbb{R} ^{\omega \times \upsilon }=[x_{1}, x_{2}, \dots, x_{\omega}]$. The original feature vectors $x_{t}$ are first standardised for each dimension and then fed through a Discrete Wavelet Transformation (DWT) to perform noise mitigation. We choose a threshold $\lambda$ of 0.7 as seen in \citep{Ye2021} which is uesd to filter away the part of the normalised time series whose frequency coefficients are higher than the threshold. The smoothed time series coming out of the DWT module is next encoded by a non-linear network $g_{enc}$ into the standardised data as $z_{t}=\mathbf{w}_{enc}x_{t}+b_{enc}$, where $\mathbf{w}_{enc} \in \mathbb{R} ^{ d \times \upsilon }, b_{enc} \in \mathbb{R} ^{d}$ are the learnable parameters from the non-linear network, and $z_{t} \in \mathbb{R}_{d}, t=0, \dots, \omega$ are the embedded vectors of dimension $d$ as inputs to the Transformer. $z_{t}$ is sensitive to the ordering of the sequence. Consequently, in order to make the Transformer aware of the sequential nature of the time series, positional encodings $z_{pos} \in \mathbb{R} ^{\omega \time \upsilon}$ are added yielding the final inputs to Transformer, $z_{t}^{'} = z_{t} + z_{pos}$.

\begin{figure}[t!]
  \centering
  \includegraphics[width=0.5\textwidth]{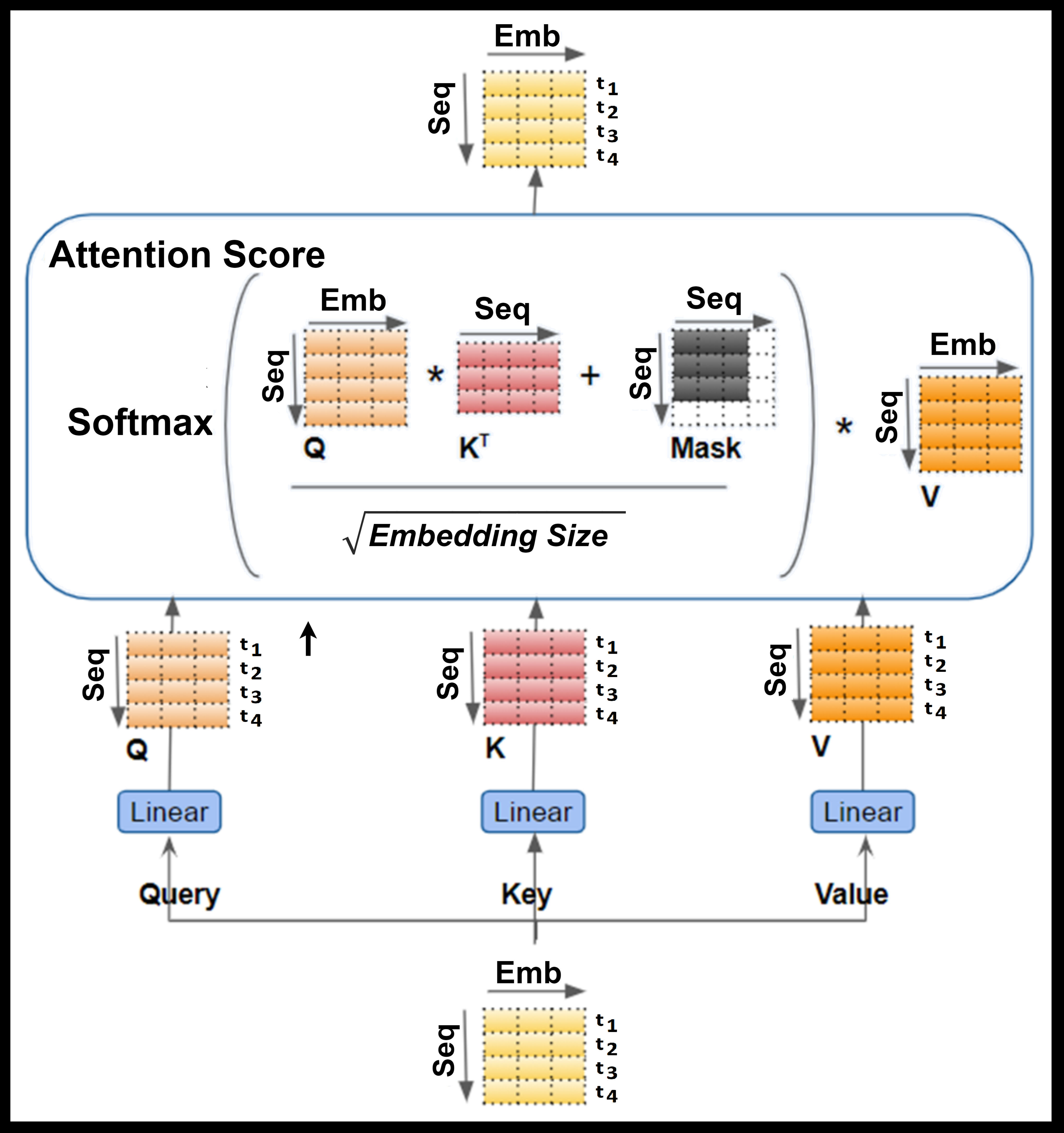}
  \caption{Structure of Transformer Encoder which is employed to produce the embedding representation of input time series $z_{t} \in \mathbb{R}_{d}, t=0, \dots, \omega$ with dimension $d$, which is the result of previously feeding minutely-frequency price and volume data through a non-linear network. Operations of the Transformer is described in section \ref{sec:encoding_of_data}}
  \label{fig:transformer}
\end{figure}

A Transformer encoder structure with multi-head self-attention is used. Through three trainable matrices (linear layers), three vectors are generated (query, key, and value) for each $z_{t}$. To compute attention, the model first compares the query vector of the current $z_{t}$ with the key vectors of $z_{\{i \neq t\}}$ at all the other times in the training sample, measuring how similar they are to each other. Next, the model calculates attention weights, which determine the importance of each $z_{\{i \neq t\}}$ in relation to the current $z_{t}$. These weights are like scores that say which data are most relevant or important for understanding the current data point. Finally, the value vectors of all the $z_{t}$ are multiplied by their corresponding attention weights and combined together, creating a new representation $c_{t}^{s}$ of the data, taking into account its context and the importance of other $z_{\{i \neq t\}}$ in understanding it. This structure and process is explained in figure \ref{fig:transformer}.

\subsection{\label{sec:method_CPC}Representation Learning of Combined Data Through CPC}

The presence of earnings data on the day they are released creates a unique context for the intra-day stock price movements, a context that CPC is perfectly placed to capture, for it not only learns to understand the context of each data point in relation to the surrounding data points (\textit{price-level context}), but it also captures the dynamics imposed by the current set of earnings data as opposed to earnings data of previous quarters (\textit{earnings-level context}). This temporal context modelling is a key feature of CPC as opposed to other forms of contrastive learning. It is about implicitly capturing essential information shared across sections of data series and is achieved by maximising what is termed as the Mutual Information (MI) between the context vector $c_{t}$ at present and a prediction vector $x_{t+k}$.

At the high level, the problem can be formulated like this: at time $t$, a set of samples $X=\left[ x_{1} ... x_{N} \right]$ is given which contains one positive sample $x_{t+k}$ and $N-1$ negative samples. The vector $x_{t+k}$ is designated as the positive sample because the goal is to maximize the implicit mutual information between $c_{t}$ and $x_{t+k}$ and minimize the mutual information between $c_{t}$ and the rest of the (negative) samples. Here $p(x_{t+k}|c_{t})$ represents the conditional probability of achieving a $x_{t+k}$ given $c_{t}$, and $k$ is the number of time steps into the future at which a prediction is to be carried out. Tackling this problem with CPC is about optimizing the InfoNCE loss which is defined in equation \ref{equ:infonce_loss}. 

\begin{equation}\label{equ:infonce_loss}
    L_{InfoNCE}(k)=-\mathbb{E}\left [ log\frac{f_{k}\left ( x_{t+k}, c_{t} \right )}{\sum_{x_{j}\in X}^{}f_{k}\left ( x_{j},c \right )} \right ]
\end{equation}Where $f_{k}$ is a scoring function. 


The loss in equation \ref{equ:infonce_loss} is the categorical cross-entropy of classifying the positive samples correctly. To capture the Mutual Information between context vector $c_{t}$ and the positive sample $x_{t+k}$, the optimal conditional probability is $p(x = t+k|X,c)$, which is defined in  \ref{equ:prob_of_pos_sample}, with $[x = t+k]$ being the indicator that sample $x_{t+k}$ is the positive sample. 

\begin{equation}\label{equ:prob_of_pos_sample}
\begin{split}
&    p(x=t+k|X,c)  \\
&   = \frac{p(x_{t+k}|c) \prod_{i=1,...,N;i\neq pos}^{}p(x_{i})}{\sum_{j=1}^{N} p(x_{j}|c) \prod_{i=1,...,N;i\neq j}^{}p(x_{i}) } \\
& = \frac{p(x_{t+k}|c)    \frac{\prod_{i=1,.,N}^{}p(x_{i})}{p(x_{t+k})}}{\sum_{j=1}^{N}\ p(x_{j}|c)    \frac{\prod_{i=1,.,N}^{}p(x_{i})}{p(x_{j})} } = \frac{\frac{p(x_{t+k}|c)}{p(x_{t+k})}}{\sum_{j=1}^{N}\frac{p(x_{j}|c)}{p(x_{i})}} 
\end{split}
\end{equation}

Where $\frac{p(x_{i}|c_{t})}{p(x_{i})}$ is the density ratio. It needs to be emphasized here that the goal of CPC is to avoid direct supervised learning for high-dimensional data, i.e. the model does not predict future observations $x_{t+k}$ directly with a generative model with probability $p(x_{t+k})$. Instead, it models a density ratio which preserves the mutual information between  $c_{t}$ and $x_{t+k}$. 

Equations \ref{equ:prob_of_pos_sample} and \ref{equ:infonce_loss} together suggest that the optimal value of the scoring function is proportional to the density ratio, i.e. $f_{k}\left ( x_{t+k}, c_{t} \right ) \propto \frac{p(x_{t+k}|c_{t})}{p(x_{t+k})}$. Optimizing the InfoNCE loss will result in $f_{k}\left ( x_{t+k}, c_{t} \right )$ estimating the density ratio. Consequently, rather than directly modelling the probability of future stock price observations which has been established to be rather difficult and expensive, CPC is chosen to do a good job of modelling a density function which preserves the mutual information between $x_{t+k}$ and $c_{t}$. 

Lastly, the CET model follows the recommendation in \citep{Oord2018} and takes the choice of modelling $f_{k}$ using the log-bilinear (LBL) model \citep{Gurav2020}. Specifically the LBL model combines context vector $c_{t}$ at time $t$ with $z_{t+k}$ at time $t+k$ (the latent representation of future stock price and volume data $x_{t+k}$) and a set of trainable weights $W_{k}$ which belongs to a linear layer in the network implementation. 

\begin{equation}\label{equ:logbilinear}
    f_{k}(x_{t+k}, c_{t})=exp\left ( z_{t+k}^{T}W_{k}c_{t} \right )
\end{equation}

With $T$ representing transposition.

This modelling arrangement effectively turns $\frac{f_{k}}{\sum_{X}^{}f_{k}}$ into a softmax function, and the final form of the InfoNCE loss function employed by the CET model is presented in equation \ref{equ:infonce_loss_full}. 

\begin{equation}\label{equ:infonce_loss_full}
\begin{split}
&    L_{InfoNCE}(k) \\
& =-\mathbb{E}\left [ 
log\frac{exp\left ( z_{t+k}^{pos}W_{k}c_{t} \right )}{exp\left ( z_{t+k}^{pos}W_{k}c_{t} \right )+\sum_{j=1}^{N-1}exp\left ( z_{t+k}^{neg}W_{k}c_{t} \right )} 
\right ]
\end{split}
\end{equation}

To wrap up this section, there are two important design choices that need to be emphasized at this stage. First, predictions are being performed for multiple future time steps, allowing the model to capture the broader structure of the minutely-frequency data. This approach goes beyond relying solely on a single step, which typically only considers the immediate smoothness of the signal \citep{Haresamudram2020b}. This results in representations that allow the model to infer a more global structure between temporally separated parts of the time-series signal, or 'slow features' as discussed at the start of this section. Second, there is a deliberate method in the selection of negative samples for the model. These samples are randomly drawn from a wide range of price data but with a specific condition: they must originate from at least five days before any earnings announcements. This precaution is taken to mitigate the risk of information leakage, which is a concern in financial markets and can lead to unfair practices like front running and insider trading. By carefully choosing negative samples that have minimal connection to the immediate post-event period, the aim is to avoid bias or data contamination, ensuring a more robust and reliable modeling process.

\subsection{\label{sec:level2}Stock movement prediction post pre-training}

Both the Transformer and Autoencoder encoders as well as the LBL model are concurrently trained employing the InfoNCE loss function. This cooperative training facilitates the prediction of the next token representation in the price sequence, conditioned upon the context $c_{t}$, by identifying concurrent patterns in price fluctuations. All of these model elements globally undergo a synchronised and comprehensive optimisation of their network weights in order to minimise the InfoNCE loss, contributing to the improvement of model performance:

\begin{equation}\label{equ:weight_minimization}
\begin{split}
& \mathbf{w} _{lbl}^{*}, \mathbf{w} _{ar}^{*}, \mathbf{w}_{enc}^{*}, \mathbf{w}_{ae}^{*} \\
& = \operatorname*{arg\,min}_{\mathbf{w}_{lbl}, \mathbf{w}_{ar}, \mathbf{w}_{enc}, \mathbf{w}_{ae}}L_{InfoNCE}(\mathbf{w}_{lbl}, \mathbf{w}_{ar}, \mathbf{w}_{enc}, \mathbf{w}_{ae}),
\end{split}
\end{equation}

where $\mathbf{w}_{lbl}^{*}, \mathbf{w}_{ar}^{*}, \mathbf{w}_{enc}^{*}, \mathbf{w}_{ae}^{*}$ represent the optimised weights for the LBL linear layer, the Transformer component $g_{ar}$, the non-linear encoding component $g_{enc}$ implemented as an MLP with one hidden layer and linear activation, as well as the three-layered (including the bottleneck layer) autoencoder $h_{ae}$ for the earnings data. 

In order for the model to predict price movement direction following the pre-training, the $c_{t}$ layer is connected to a new output layer that can captures the three movements: up, down and hold. $\mathbf{w}_{ar}^{*}$, $\mathbf{w}_{enc}^{*}$ and $\mathbf{w}_{ae}^{*}$ obtained from the last step are all fixed, and weights of the output layer are learnt from scratch using cross-entropy loss with a softmax activation function and Adam optimizer \citep{Kingma2014} with a learning rate of 2e-4 as seen in \ref{equ:cross_entropy_softmax}. Note the LBL layer is not required in the price prediction operations as its presence was needed entirely to optimize the InfoNCE loss during unsupervised pre-training.

\begin{equation}\label{equ:cross_entropy_softmax}
\begin{aligned}
L=-\sum_{i}y_{i}log(p_{i})  
\text{, where } p_{i}=\frac{e^{c_{i}}}{\sum_{k=1}^{N}e_{k}^{c}}
\end{aligned}
\end{equation}

\section{\label{sec:experimental_results_and_analysis}Experiments, Results and Analysis}

The main objective of this research is to assess the effectiveness of CPC-based self-supervised representation learning in capturing the impact of earnings shocks on intra-day stock price movements for intra-day algorithmic trading and gain insights into the underlying mechanisms. To accomplish these goals, we have designed a series of progressive experiments that provide increasingly insightful observations and analysis into these aspects. In all these experiments, we compare the performance obtained by CPC against the following list of relevant/state-of-the-art unsupervised approaches, each performing their respective pretext tasks:

\begin{description}
  
  \item[\small $\bullet$ Autoencoder (AE)] {\small Autoencoders are respectively applied to earnings data and stock price data generating $c_{t}^{e}$ and $c_{t}^{s}$.}
  
  \item[\small $\bullet$ Transformer with Masked Language Modelling (MLM)] {\small MLM is a self-supervised learning technique used in the groundbreaking NLP model, BERT \citep{devlin2019}. When applying MLM to the Transformer, we randomly mask sections of stock price data sequence and predict the masked sections from the unmasked ones. To make MLM work with numerical sequence, we substitute the original softmax cross entropy loss (designed for word token prediction) with a more appropriate Mean Squared Error (MSE) regression loss.}

  \item[\small $\bullet$ SimCLR] {\small SimCLR \citep{Chen2020} is another state-of-the-art self-supervised contrastive learning model. Although it was originally conceived for image data, it has now been adapted to handle the price data sequences with the Transformer. To generate positive samples, variations are infused into each input data sequence by adding random Gaussian noise, which emulates the data augmentation techniques used in the original image data-based design. The subsequent operations, embedding, similarity calculation, and loss optimization, remain unchanged.}
  
\end{description}

For a vertical comparison, three supervised baselines have also been created for benchmark. These baselines are constructed by incrementally removing key elements from the CET model design:

\begin{description}
    
  \item[\small $\bullet$ SupRep] {\small Direct supervised learning using the context vector $c_{t}$. This represents the same network structure used in CET with no CPC-based pre-training and hence without the InfoNCE loss.}
  
  \item[\small $\bullet$ SupRaw] {\small Direct supervised learning with minimum representation learning (and hence noted `raw'). In this model, price and volume data feed directly to the Transformer as CET. Financial metrics from earnings report are not encoded and are concatenated to the Transformer output directly.} 
  
  \item[\small $\bullet$ SupRaw2] {\small Direct supervised learning with the Transformer encoder on price and volume data only and no financial metrics.  }
  
\end{description}

\subsection{\label{sec:data_setup}Data Setup}
We have selected a diverse range of data from \citep{Ye2021}, as outlined in table \ref{tab:EarningsFeatureList} that are considered to be most indicative of a company's financial soundness. This dataset is acquired from Bloomberg and subjected to pre-processing procedures to address missing and outlier data. Additionally, the data is adjusted for dividends and stock splits, converted to ratio format when necessary, and normalised using the Z-Score method. In addition, we also include each quarter’s Earnings Surprise (reported EPS minus market estimated EPS) and the change between current quarter's Earnings Surprise and that of the previous quarter. In total, we have selected 38 financial metrics that collectively reflect a company's earnings quality for each quarter.

The minutely-frequency price and volume data are programmatically downloaded through the public API of the Investors Exchange (IEX) \citep{IEX} which is a national US stock exchange that also supplies historically traded OHLCV (Open, High, Low, Closed price and Volume) data that are publicly accessible for benchmarking purposes. There are 390 minute data points on each trading day and linear interpolation is employed to address rare missing data problems. To ensure equal representation and optimal price liquidity, we choose companies from the S\&P500 index and divide them into the nine industrial sectors categorised by Bloomberg: \textit{Basic Materials, Communications, Consumer Cyclical, Consumer Non-Cyclical, Energy, Financial, Industrial, Technology, and Utilities}, and select 10 companies from each sector to form our company universe as seen in table \ref{tab:companies_selected_for_CET_experiments}. We retrieve minutely-frequency data for the five business days after each company's quarterly financial reporting, or from the announcement date if it occurs before the market opens. Data from the days prior to financial results announcements are also required and used as negative samples as described in section \ref{sec:method_CPC} but are not part of the training/testing data set. We have intentionally selected a window of 10 action-packed financial quarters, starting from the fourth earnings quarter in 2020 (around October) through the first quarter of 2023 (around January). Within this timeframe, the US stock markets initially witnessed extended periods of upward momentum, primarily driven by the widespread availability of Covid-related funds to retail investors, immediately followed by prolonged periods of decline, attributed to concerns over inflation and the ongoing war in Ukraine.

\begin{table}[ht]
\centering
\begin{tabular}{|p{4.0cm}|p{3.7cm}|}
\hline
Cash change \%   & Operating Margin                           \\ \hline
Operating Cash change \%                   & Price to Book Ratios                            \\ \hline
Cost of Revenue change \%               & Price to Cashflow Ratios                  \\ \hline
Current Ratio                     & Price to Sales Ratios                   \\ \hline
Dividend Payout Ratio & Quick Ratio              \\ \hline
Dividend Yield               & Return On Assets                      \\ \hline
Free Cash Flow change \%              & Return On Common Equity                       \\ \hline
Gross Profit change \%                & Revenue change \%                 \\ \hline
Operating Income change \%                 & Short Term Debt change \%                          \\ \hline
Inventory Turnover change \%                       & Total Liabilities change \%                 \\ \hline
Net Debt to EBIT    & Total Asset change \% \\ \hline
Net Income change \%               & Total Debt to Total Assets                        \\ \hline
Operating Expenses change \%               & Total Debt to Total Equity                        \\ \hline
Operating Income change \%               & Total Inventory change \%                       \\ \hline
               
\end{tabular}
\caption{\label{tab:EarningsFeatureList}Representative earnings report metrics chosen as input features. Data which are not reported as ratios in the quarterly filings with Securities and Exchange Commission (SEC) have been converted to percentage change (\%). Data of each variable have been standardised over all training data sets, statistics of which are then used to standardise the testing data set.}
\end{table}

\subsection{\label{sec:a_look_at_pre_training}A Look At Pre-training}

The initial phase of the entire learning process involves CPC-based self-supervised pre-training, where the weights obtained are later used as a feature extractor in algorithmic trading activities. To evaluate how effectively these representations have been learned, model performances are analyzed using a classifier network tasked with predicting stock price movements. This analysis focuses specifically on the data from the day immediately following the release of earnings reports (day 1), a strategic choice made to ensure the model performances are measured under the most pronounced effects of these financial events. As a result of this methodology, the dataset used includes 900 days of financial earnings data, complemented by intra-day minute-by-minute price and volume information. This extensive dataset serves as the foundation for pre-training the four self-supervised models.

After the unsupervised pre-training phase is completed, the network is augmented with an additional fully connected layer that serves as a task-specific component. This layer utilizes softmax cross-entropy loss for its computations. Next, to refine the pre-trained models, three identical but separate experiments are conducted using 20\%, 50\%, and 80\% of the entire dataset, selected at random for fine-tuning. Additionally, another 20\% of the dataset, chosen randomly, is reserved for testing purposes. In each experiment the benchmark supervised learning models also consume the same sets of training and testing data.

There are two important aspects to highlight here regarding experiment setup. Firstly, for the self-supervised models, both the pre-trained segments of the network and the newly incorporated output layer undergo fine-tuning. Secondly, the model employs an Adam optimizer with a learning rate of 2e-3 during the pre-training stage, which is then reduced to 2e-4 during the fine-tuning phase. This reduction in the learning rate is strategic; a too high learning rate during fine-tuning might cause the model to neglect the general features it learned during the pre-training stage.

\begin{table*}[]
\resizebox{1.0\textwidth}{!}{%
\begin{tabular}{c|c|c|c|c|c|c|c|}
\cline{2-8}
                                      & CET                    & AE            & MLM           & SimCLR        & SupRep        & SupRaw                 & SupRaw2       \\ \hline
\multicolumn{1}{|l|}{20\% of dataset} & \textbf{57.74 + 0.300} & 55.05 + 0.364 & 54.78 + 0.456 & 56.37 + 0.346 & 55.01 + 0.789 & 55.52 + 0.825          & 54.45 + 0.855 \\ \hline
\multicolumn{1}{|l|}{50\% of dataset} & 58.50 + 0.296          & 56.62 + 0.308 & 56.18 + 0.343 & 57.67 + 0.458 & 58.88 + 0.637 & \textbf{59.35 + 0.687} & 55.88 + 0.658 \\ \hline
\multicolumn{1}{|l|}{80\% of dataset} & 58.01 + 0.268          & 56.69 + 0.274 & 55.60 + 0.273 & 57.00 + 0.487 & 58.26 + 0.767 & \textbf{59.72 + 0.805} & 55.50 + 0.722 \\ \hline
\end{tabular}}
\caption{Prediction success rate (\%) on stock movements using earnings data and minutely-frequency stock time series data, comparing performance of models that have been pre-trained and fine-tuned against those of straight supervised learning benchmark models. For self-supervised models, while the entire data set is used for pre-training, 20\%, 50\%, and 80\% of the overall dataset are separately used for fine-tuning. For supervised learning models, the same portion of the overall data set is used for direct supervised learning. Each cell has a mean success rate (\%) and a standard deviation.}
\label{tab:pre_training_results}
\end{table*}

Table \ref{tab:pre_training_results} presents the classification results from the experiments with each cell showing the mean classification rate and standard deviation. Here a few observations are made:

First and foremost, CET emerges as the leading self-supervised learning model, beating its benchmark peers in this category. The less favorable performance of MLM underlines the inherent challenges of applying context-based pre-training methods to numerical data, which lacks the same kind of contextual relationships as language data. Meanwhile, SimCLR's performance, being close to CET, illustrates the robust potential of contrastive pre-training from another perspective.

Second, in the first scenario where label information is limited due to only 20\% of data being used for fine-tuning/direct supervised learning, pre-trained models expectedly take the upper hand against supervised models. This is a good example that showcases their strength in leveraging the unlabeled data for pre-learning useful representations, allowing models to quickly converge to an equilibrium state with comparatively smaller amount of data. The benefit of pre-training, although diminishes with increasing training data, is evident. 


Third, while supervised models initially lag behind their pre-trained counterparts when training data is scarce, they are able to narrow or even close the performance gap with the introduction of more training data. However, it is observed that the supervised model performances appear to diminish as the dataset size becomes quite large, which the author considers as a classic case of overfitting. The same is in fact also observed with the fine-tuning of self-supervised models. 

Fourth, it can also be observed that the SupRaw model surpasses its fellow supervised learning models, including SupRep. This observation is intriguing, and it could potentially be attributed to the additional complexity in SupRep's network structure, originally intended for unsupervised pre-training within the CET model framework. Its excessive sophistication appears to have undermined its performance compared to a more simplified network when used in a direct supervised learning setting.

Fifth, on the other hand, the substantial drop in the performance of SupRaw2 (where earnings data is excluded) underlines the vital role of this data in the classification task, a challenge that pre-trained models might handle better due to their ability to extract useful information from diverse and comprehensive datasets.

Lastly, the high standard deviation in performance among all supervised learning models points towards their relative instability. In contrast, models that undergo unsupervised pre-training exhibit consistency and stability, reinforcing the value of pre-training in achieving reliable performances.

\subsection{\label{sec:level2}Task-specific Training}

The value of fine-tuning a pre-trained model on a downstream task is most apparent when there's additional labelled data that is specific to the downstream task. This allows the model to adjust its parameters to better fit the specific task. To evaluate the value of this feature, we take the self-supervised models that were pre-trained and fine-tuned in the previous section, and respectively train them to work with data from companies in each of the nine industrial sectors defined in section \ref{sec:data_setup}.

Specifically for the CET model, there is firstly the pre-training step using CPC. Next the model is fine-tuned with 60\% of all company data, after which the learned weights $\mathbf{w}_{ar}^{*}, \mathbf{w}_{enc}^{*}, \mathbf{w}_{ae}^{*}$ from section \ref{sec:method_CPC} are frozen and used in the classifier network. Sector-specific company data from half of the remaining labeled data is used to train the classifier network using cross-entropy loss. Essentially for the self-supervised models, the whole data set is split into three portions for fine-tuning / sector specific supervised training / sector specific testing in a 60:20:20 ratio, with the supervised training and testing data being sector-specific. For the benchmark supervised models, the first 60\% of data from all companies, in conjunction with sector-specific company data in half of the remaining data, is used for direct supervised training. Sector-specific data in the final 20\% of the whole dataset is reserved for testing.

\begin{table*}[]
\resizebox{1.0\textwidth}{!}{%
\begin{tabular}{|l|l|l|l|l|l|l|l|}
\hline
\textbf{Sector}       & \textbf{CET}          & \textbf{AE}  & \textbf{MLM} & \textbf{SimCLR} & \textbf{SupRep} & \textbf{SupRaw}       & \textbf{SupRaw2} \\ \hline
Basic Materials       & \textbf{60.4 + 0.260} & 58.0 + 0.275 & 59.1 + 0.250 & 60.2 + 0.255    & 58.5 + 0.530    & 59.0 + 0.545          & 54.9 + 0.560     \\ \hline
Communications        & \textbf{57.5 + 0.210} & 55.0 + 0.305 & 56.0 + 0.280 & 57.3 + 0.288    & 56.9 + 0.610    & 57.0 + 0.625 & 52.5 + 0.640     \\ \hline
Consumer Cyclical     & 60.1 + 0.253          & 59.5 + 0.258 & 57.5 + 0.242 & 59.9 + 0.248    & 58.7 + 0.508    & \textbf{60.3 + 0.523} & 55.5 + 0.538     \\ \hline
Consumer Non-Cyclical & \textbf{61.4 + 0.245} & 59.5 + 0.253 & 60.5 + 0.237 & 61.1 + 0.243    & 59.9 + 0.498    & 60.2 + 0.512          & 56.9 + 0.528     \\ \hline
Energy                & \textbf{55.2 + 0.315} & 54.8 + 0.328 & 52.5 + 0.303 & 55.0 + 0.310    & 53.3 + 0.650    & 53.8 + 0.665          & 50.5 + 0.680     \\ \hline
Financial             & 58.6 + 0.278          & 57.8 + 0.292 & 55.8 + 0.268 & 58.5 + 0.276    & 56.9 + 0.580    & 57.2 + 0.598          & 53.8 + 0.615     \\ \hline
Industrial            & \textbf{62.3 + 0.238} & 60.4 + 0.246 & 59.5 + 0.232 & 62.1 + 0.238    & 61.0 + 0.485    & 61.5 + 0.500          & 57.8 + 0.515     \\ \hline
Technology            & \textbf{59.3 + 0.267} & 58.4 + 0.280 & 57.8 + 0.259 & 59.1 + 0.263    & 58.0 + 0.565    & 58.5 + 0.580          & 58.0 + 0.596     \\ \hline
Utilities             & \textbf{51.0 + 0.240} & 50.5 + 0.355 & 48.5 + 0.325 & 50.8 + 0.333    & 48.8 + 0.698    & 49.2 + 0.710          & 46.5 + 0.730     \\ \hline
\end{tabular}}
\caption{Prediction success rate (\%) on stock movements using earnings data and minutely-frequency stock time series data. After pre-training, self-supervised models are fine-tuned with 60\% of all company data. The classifier layer is trained with sector-specific data once fine-tuned weights have been frozen. Data is split in 60:20:20 ratio for fine-tuning, sector specific supervised training, and testing. Each cell has a mean success rate (\%) and a standard deviation.}
\label{tab:results_section_5.2}
\end{table*}

As seen in the result table \ref{tab:results_section_5.2}, the CET model emerges as a consistent top performer across most sectors. Its success rates are generally on the higher end, indicating its robustness and ability to generalize. The relatively small deviations in its success rates across sectors further emphasize its adaptability. In essence, CET's performance, as predicted, underscores its superior generalization capabilities. This appears particularly true when working with such diverse set of stock-related data irrespective of sector-specific nuances, especially when compared to other models that tend to show a wider range of outcomes depending on the industry in focus. For instance, MLM, designed on the principles of NLP's BERT model, reflects varied success rates, indicating its potential struggles with sector-specific nuances. SimCLR, which is adapted from image data-based design to the current context, portrays results close to CET but varies more distinctly between sectors. 

Among the supervised models, SupRep and SupRaw once again present intriguing performances similar to what was seen in the last experiment \ref{sec:a_look_at_pre_training}. SupRep, which employs direct learning from the context vector, excels in specific sectors but appears to lack consistency in its performance. SupRaw, with its minimal representation learning, showcases results that, in certain sectors, even surpass SupRep. This could be indicative of the potency of raw data in making accurate predictions. Models leveraging earnings data tend to fare better, aligned with the expectation that earnings data provide crucial insights into a company's financial health and prospects, often serving as an indicator of future stock performance. As such, these models are equipped with richer context, leading to more accurate predictions.  In contrast, SupRaw2 which lacks the earnings data inputs, predictably has a lower performance than its counterparts.

\subsection{\label{sec:level2}Preserving Predictive Power of Earnings Data}

The previous experiments consider stock movements only on the first day after the release of earnings data (or the same day if release was made before market open), denoted as $D_{1}$. While the benefits of having earnings data is most profound at this point, this daily trading is highly volatile and the data's predictive power rapidly comes down after Day 1 \citep{Ye2021}. Consequently, further test is arranged in this section to extend this notion of locality to one week (i.e., five business days) and evaluate how well the CET model pre-trained with Day 1 data will preserve the predictive power of earnings data. Specifically, after pre-training on $D_{1}$, the less volatile near-term performances are evaluated over the next four business days, $D_{2}$ - $D_{5}$. On each day, models would receive 60\% of the stock and earnings data for training, while the remaining 40\% are employed as the test set.

\begin{table*}[]
\resizebox{0.6\textwidth}{!}{%
\begin{tabular}{|c|c|c|c|c|c|c|c|}
\hline
\textbf{Model} & \textbf{CET} & \textbf{AE} & \textbf{MLM} & \textbf{SimCLR} & \textbf{SupRep} & \textbf{SupRaw} & \textbf{SupRaw2} \\ \hline
\textbf{Day 2} & 58.52      & 56.87     & 56.05      & 57.15         & 56.50         & 57.25         & 54.75          \\ \hline
\textbf{Day 3} & 58.31      & 56.55     & 56.15      & 57.03         & 55.40         & 56.10         & 54.60          \\ \hline
\textbf{Day 4} & 58.02      & 56.52     & 56.10      & 56.89         & 54.80         & 55.50         & 54.02          \\ \hline
\textbf{Day 5} & 58.04      & 56.51     & 56.08      & 56.77         & 53.72         & 54.90         & 54.05          \\ \hline
\end{tabular}}
\caption{Comparing how CET and the benchmark models perform in the days following the release of a company's financial earnings data. For self-supervised learning models, all the Day 1 data is used for pre-training and fine tuning. The data on each of the subsequent 4 days are used for continuous fine-tuning of the models. For supervised learning models, they are re-trained everyday using the new day's data.}
\label{tab:results_section_5.3}
\end{table*}

\begin{figure*}[t!]
  \centering
  \includegraphics[width=0.5\textwidth]{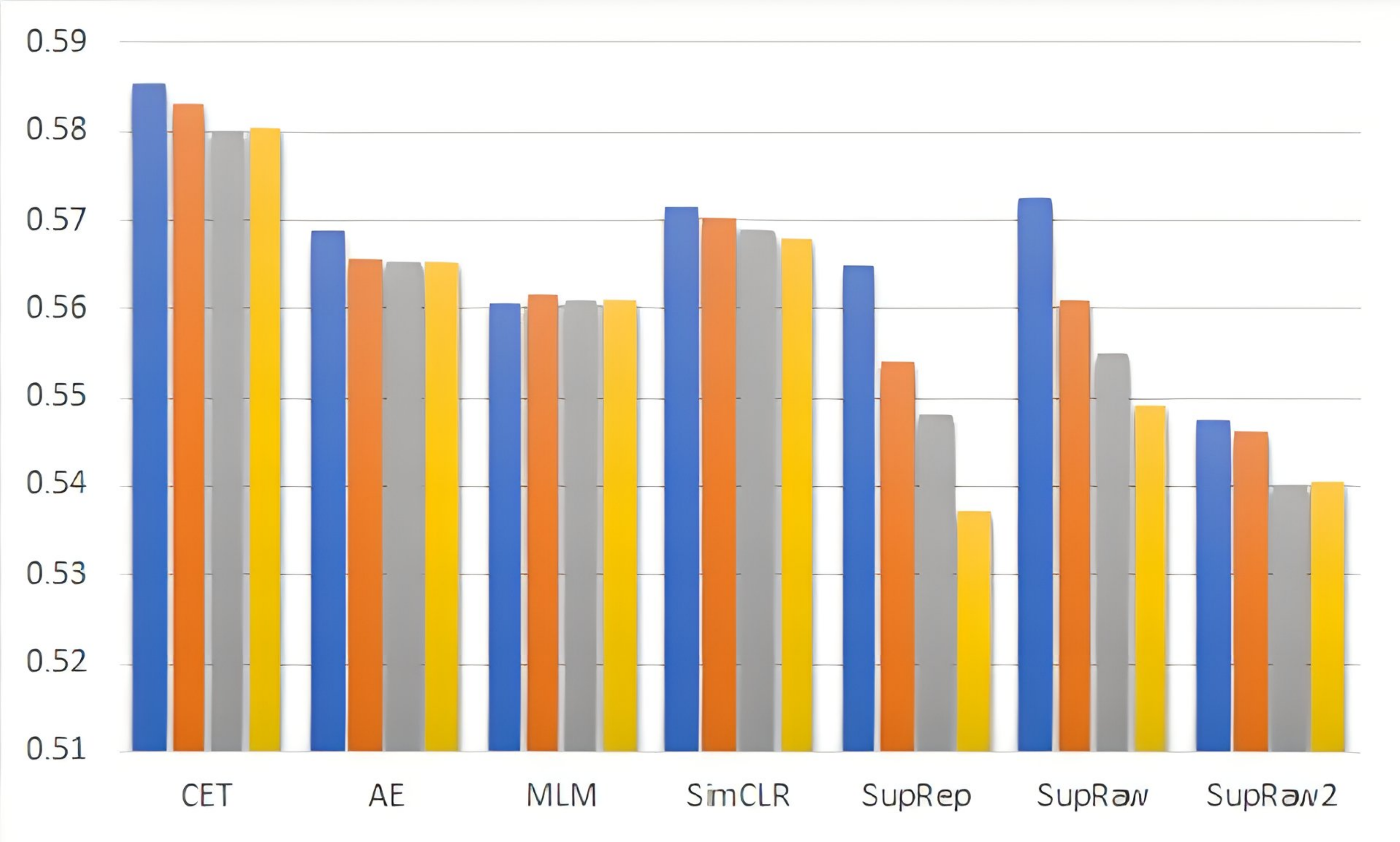}
  \caption{Chart representation of data in table \ref{tab:results_section_5.3} portraying varying model performances over the 4 days since the release of earnings data. The y-axis is prediction success rate. For each model, the four bars going left to right represent day 2 -- day 5.}
  \label{fig:4day_Model_performance}
\end{figure*} 

As seen in the table \ref{tab:results_section_5.3} and figure \ref{fig:4day_Model_performance}, the CET model showcases a consistency throughout the four-day period. Beginning at 58.52\% on Day 2, it slightly drops to 58.02\% by Day 4 but then stabilises and even marginally improves to 58.04\% by Day 5. This subtly upward trend between Day 4 and Day 5 could be due to different stock data characteristics on these days and appears unique to CET, underscoring its robustness and its ability to navigate the complexities of the stock data landscape. It is this kind of nuanced learning that makes CET stand out as it appears to course-correct its understanding from the previous day's downturn.

In contrast, the Autoencoder (AE) and Transformer with MLM models exhibit stable performances over the period but do not manage to eclipse CET's success rates. Both models' performance are particularly consistent, with MLM's performance sees a slight increase on Day 3, followed by a minor decrease over the next two days, suggesting a degree of sensitivity to temporal patterns in the data.

SimCLR presents a noticeable continuous decline from Day 2 to Day 5. Starting at 57.15\% on Day 2, it slides down to 56.77\% by Day 5. This progressive decline might hint at SimCLR's challenges in adjusting to the diminishing relevance of earnings data over subsequent days, a trait that diverges from its general self-supervised counterparts.

Among the supervised models, SupRep distinctly portrays a sharp diminishing trend from 56.50\% on Day 2 to 53.72\% by Day 5. This trajectory reiterates the notion of supervised models being more reliant on fresh earnings data, and as its immediate impact diminishes, so does their prediction capability. The model might also subject to over-complexity of its network. SupRaw also reveals a similar declining trend, although its starting point on Day 2 is higher than that of SupRep, something that we also observed in the previous experiment. Meanwhile, SupRaw2, contrary to SupRep, remains almost consistent, with a tiny dip between Day 2 and Day 4, but exhibits a recovery on Day 5. 

This experiment highlighted two key findings. First, the ability of self-supervised models, particularly CET, to recalibrate and learn from evolving patterns in financial datasets is remarkable. Their performance trajectory indicated not just learning but a nuanced understanding of changing data dynamics. Second, while supervised models can be powerful predictors, their rigidity in the face of rapidly altering feature significance, like earnings data, can be a limitation. As model performances vary over time, it's evident how critical fresh earnings data is to predictions. As this data ages, models' strengths and weaknesses become apparent, highlighting differences in their design and learning methods.

\subsection{\label{sec:level2}Ablation Analysis}

One of the key model features of CPC is its ability to make use of positive samples from multiple time steps (latent steps) into the future. It is important to evaluate the impact of changing the latent step size on the result of unsupervised pre-training which is what this study attempts to explore. The natural first step we have taken here is to examine the InfoNCE losses. Meanwhile, due to the predictive nature of CPC, we also assess the similarity between the predicted context vectors and the corresponding positive sample vectors. This comparison aims to evaluate how closely the model's predictions align with the actual future data and is measured using the cosine similarity score of two vectors $\mathbf{A}$ and $\mathbf{B}$, $\frac{\mathbf{A} \cdot \mathbf{B}}{\|\mathbf{A}\| \|\mathbf{B}\|}$. Both metrics are assessed across varying latent step sizes, ranging from 1 to 20. This assessment follows the experimental procedures outlined in section \ref{sec:a_look_at_pre_training} `A Look at Pre-Training'. For each step size, the study records the equilibrium loss rate and calculates the average similarity score. 

\begin{figure*}[t!]
  \centering
  \includegraphics[width=0.6\textwidth]{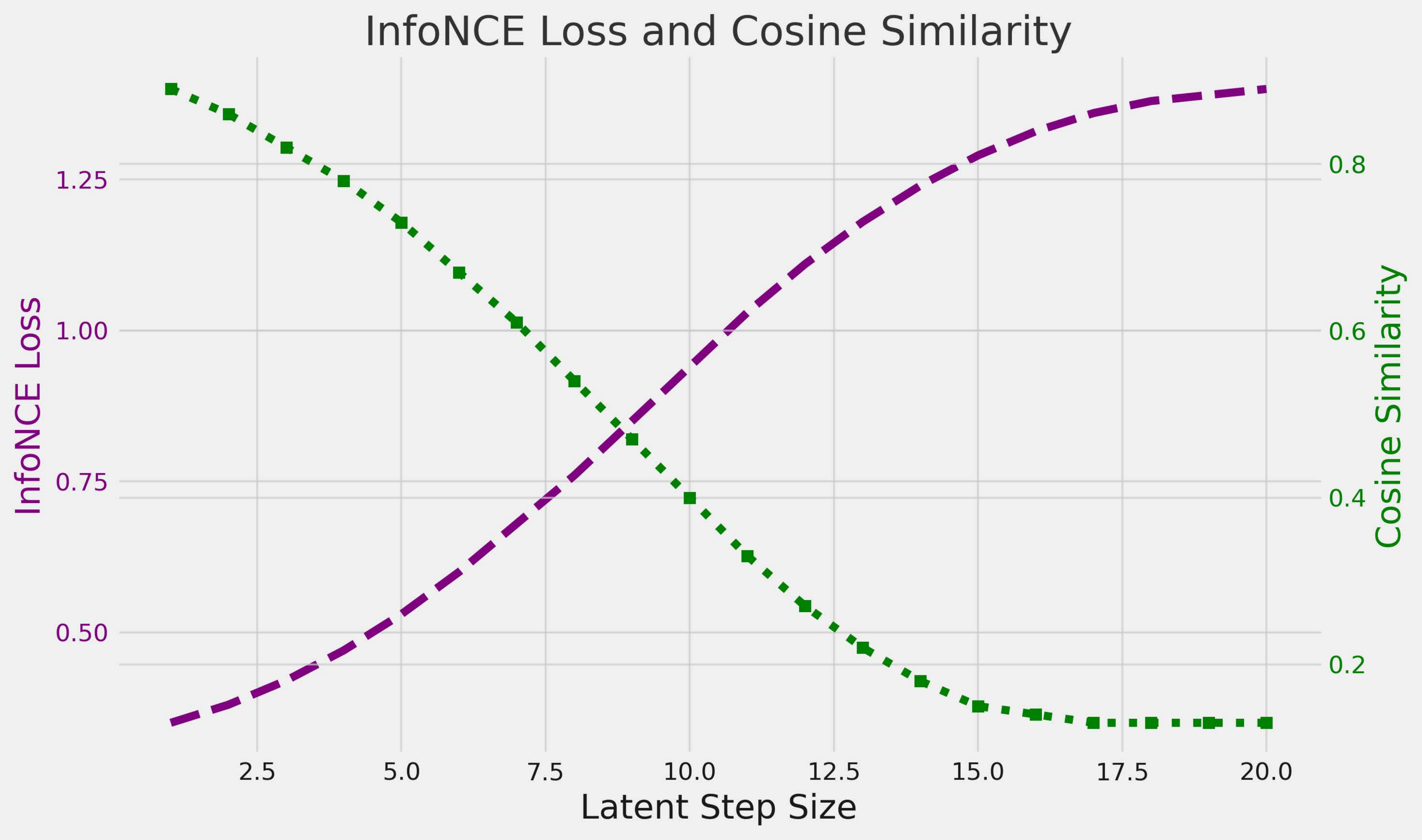}
  \caption{Ablation study results portraying the impacts by a changing latent step size on the InfoNCE loss and similarity score between predicted context vector and the corresponding positive sample vector}
  \label{fig:InfoNCE_loss_and_similarity_score}
\end{figure*}

The results of both evaluations are presented in figure \ref{fig:InfoNCE_loss_and_similarity_score} which offers a wide-ranged view of the CPC model's performance from the angle of traditional study of loss and predictability. The result on InfoNCE loss demonstrates that the model is highly effective in short-term predictions but struggles with longer prediction horizons. This is corroborated by the decreasing cosine similarity for longer prediction steps highlighting the increasing divergence between the model's predictions on the context vector and the benchmark positive vector. 

\begin{figure*}[t!]
  \centering
  \includegraphics[width=0.6\textwidth]{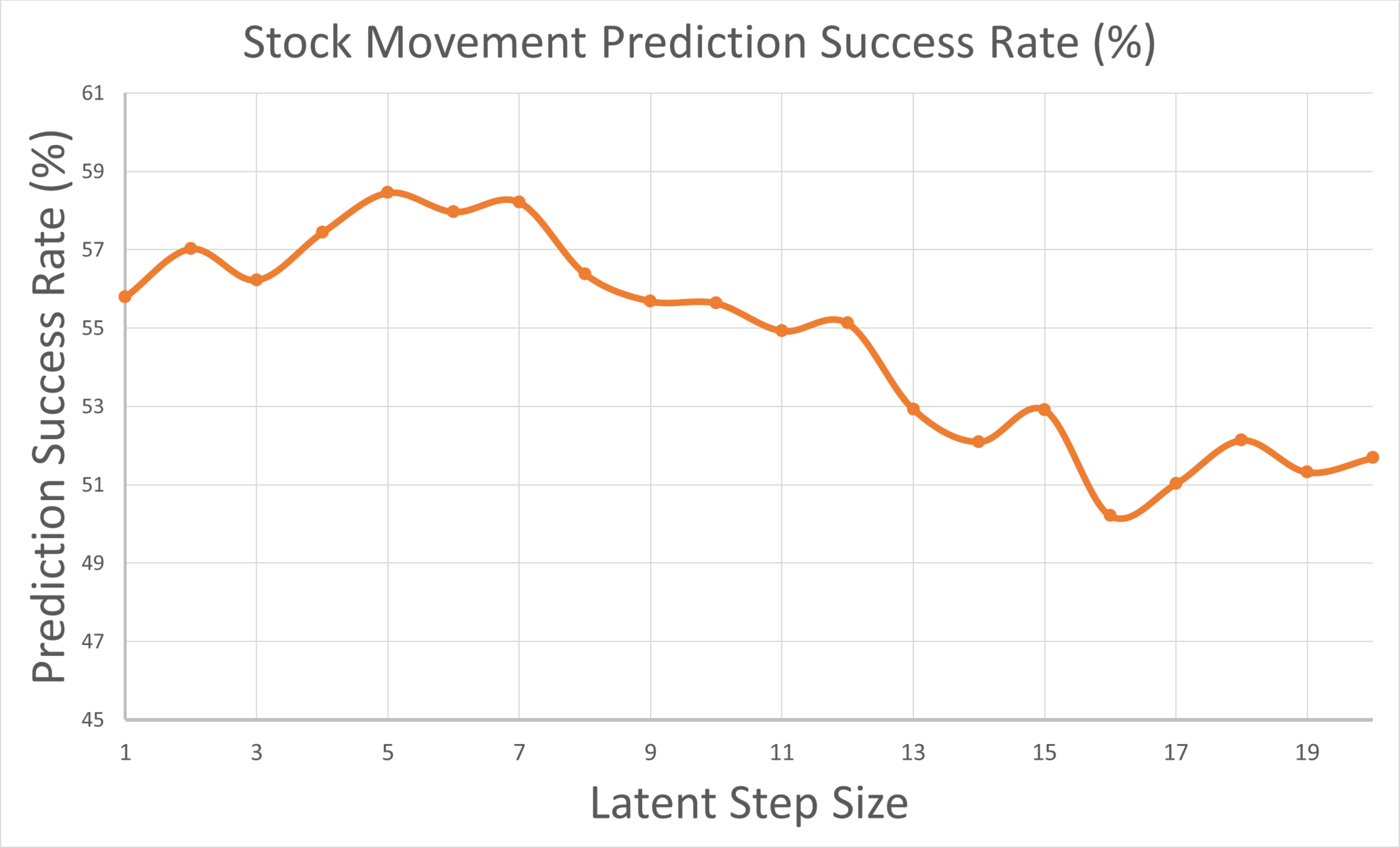}
  \caption{Ablation study results portraying the impacts by a changing latent step size on the InfoNCE loss and similarity score between predicted context vector and the corresponding positive sample vector}
  \label{fig:Ablation_analysis}
\end{figure*}

With CPC's key characteristic being its use of contrastive training, it is understood that achieving the lowest InfoNCE loss or the highest cosine similarity score does not necessarily translate to the most effective or practical model performance. It is possible that a model predicting just one future step might score well in these metrics but fail to recognise the longer-term patterns critical in financial time series data. Moving beyond initial assessments focused solely on unsupervised pre-training, this experiment next explores the model model after fine-tuning. This examination follows the same experimental setup detailed in the section \ref{sec:a_look_at_pre_training}, while using 50\% of all data to fine-tune the CET model. Again, using the same evaluation approach as seen in that section, success rate of predicting stock movements is measured for each of the latent step sizes. The outcome of this experiment is detailed in figure \ref{fig:Ablation_analysis} which shows that, while the prediction success rate fluctuates a lot, it initially goes up as data 'further into the future' is used and stays at an elevated level until after step size of 7. This observation seems to support the earlier theory that the main benefit of this kind of training is to create strong features that help with future predictions, regardless of how the InfoNCE loss is measured on its own. 

The two sets of results seen in figure \ref{fig:InfoNCE_loss_and_similarity_score} and \ref{fig:Ablation_analysis} complement each other and highlight CPC's core principle: balancing immediate precision, indicated by lower loss or higher similarity in short-term predictions, against the ability to identify complex, long-term trends. A model excelling in short-term forecasting might score well in certain metrics, yet it could lack the robustness and broader applicability that our model aims for. Our CET model is designed to predict over extended periods. This approach of forecasting further ahead can enable the model to grasp more abstract, significant aspects of the data, which are vital for its adaptability and wider use.

\section{Conclusion}
This research, centred on the predictive potential of earnings data, offers crucial insights into the current capabilities of stock prediction models. Earnings data, usually a key signal for financial analysts, has a strong impact right after it's released. But our findings show that its ability to predict quickly decreases, highlighting how fast-changing and transient this information can be.

The introduction of the CPC and Transformer-powered Contrastive Earnings Transformer (CET) model marks a major step forward in this area. CET, unlike its contemporaries, showcases a consistent performance across varied industrial sectors and, crucially, manages to uphold the predictive relevance of earnings data over a more extended period. While other models, especially the supervised ones, struggled as the impact of earnings data diminished over time, CET remained effective. It not only maintained a steady prediction rate, but also demonstrated an ability to recalibrate based on evolving stock data patterns. In the ever-changing world of stock markets, where the release of earnings data is a significant event, the ability to adapt to shifting data trends in the days after the release becomes especially crucial.

In contrast, traditional supervised models, while showing promise in specific instances, struggled to maintain their efficacy as the earnings data aged. Their declining performance over days highlighted the potential limitations they might have in real-world trading scenarios. As we explore the intricate dynamics of stock predictions, the CET model distinguishes itself with its adaptability and expertise in leveraging earnings data, offering a promising avenue for trading through earnings seasons. 

That being said, although it hadn't had much of an impact on the results of this study, noise in stock data can always have a major impact on machine learning models. The impact on this research would have been more pronounced if test data had not been intentionally sourced from a period with significant market volatility. One focus of future studies in this area should be more on exploring how well the CET model behaves under various market conditions and how it is influenced by different hyperparameters. Separately, the effectiveness of learning often depends on the selection of negative pairs, particularly if the dataset has underlying similarities across different sequences or time windows. This issue requires sophisticated negative sampling strategies to maintain stable learning efficacy over time, which in itself is a major future research direction. Lastly, as seen in section \ref{sec:a_look_at_pre_training}, high network complexity can play hindrance to future up-scaling of the model and this is an area requiring future optimization. Nevertheless, as the complexities of the financial domain expand, we hope the CET model would emerge as a promising benchmark model for future advancements in stock market predictions.


\section{Appendix}
Table \ref{tab:companies_selected_for_CET_experiments} provides the stock symbols of all the companies selected for testing the CET model and the chosen benchmarks. 

\begin{table*}[]
\centering
\resizebox{0.9\textwidth}{!}{%
\begin{tabular}{|c|c|c|c|c|c|c|c|c|}
\hline
\textbf{Basic Materials} & \textbf{Communications} & \textbf{Consumer, Cyclical} & \textbf{Consumer, Non-cyclical} & \textbf{Energy} & \textbf{Financial} & \textbf{Industrial} & \textbf{Technology} & \textbf{Utilities} \\ \hline
APD                      & AMZN                    & AN                          & ABT                             & APC             & AFL                & CAT                 & EA                  & AES                \\ \hline
AA                       & T                       & AZO                         & AET                             & APA             & ALL                & CBE                 & EMC                 & AEE                \\ \hline
ATI                      & CBS                     & RL                          & CAH                             & BHI             & AXP                & CSX                 & FIS                 & AEP                \\ \hline
CF                       & CTL                     & BBY                         & AGN                             & COG             & AIG                & CMI                 & FISV                & CNP                \\ \hline
DOW                      & CSCO                    & BWA                         & MO                              & CHK             & AMT                & DE                  & HPQ                 & CMS                \\ \hline
DD                       & CMCSA                   & KMX                         & ABC                             & CVX             & AMP                & DOV                 & INTC                & ED                 \\ \hline
EMN                      & GLW                     & CCL                         & HRB                             & COP             & AON                & ETN                 & IBM                 & D                  \\ \hline
FMC                      & DISCA                   & M                           & ADM                             & CNX             & AIV                & EMR                 & INTU                & DTE                \\ \hline
FCX                      & EBAY                    & COST                        & ADP                             & DVN             & AIZ                & MMM                 & MU                  & DUK                \\ \hline
IFF                      & EXPE                    & DHI                         & AVY                             & EOG             & AVB                & FDX                 & MSFT                & EIX                \\ \hline
\end{tabular}}
\caption{The CET model is tested using ten S\&P500 companies from each of the Bloomberg industrial sectors, and the stock symbols for the chosen companies are provided here.}
\label{tab:companies_selected_for_CET_experiments}
\end{table*}

Table \ref{tab:CET_hyperparameters} lists all the key hyperparameters of the CET model. 

\begin{table*}[]
\centering
\begin{tabular}{|l|l|}
\hline
Time series length                                           & 50                                                \\ \hline
Size of latent representation by the price encoder $g_{enc}$ & 128                                               \\ \hline
Number of hidden layers in price encoder                     & 1                                                 \\ \hline
Transformer hidden dimention                                 & 128                                               \\ \hline
Earnings data autoencoder context vector dimension           & 128                                               \\ \hline
Number of negative samples                                   & 20                                                \\ \hline
Number of forward steps                                      & 5                                                 \\ \hline
Learning rate (pre-training)                                 & 0.001                                             \\ \hline
Learning rate (fine tuning and regular supervised learning)  & 0.0001                                            \\ \hline
Batch size                                                   & 64                                                \\ \hline
Optimiser                                                    & Adam optimiser with   standard default parameters \\ \hline
\end{tabular}
\caption{Key hyperparameters of the CET model.}
\label{tab:CET_hyperparameters}
\end{table*}

\bibliographystyle{apacite}

\bibliography{cas-refs}


\end{document}